\documentclass[10pt,twocolumn,letterpaper]{article}

\usepackage{wacv}
\usepackage{times}
\usepackage{epsfig}
\usepackage{graphicx}
\usepackage{amsmath}
\usepackage{amssymb}
\usepackage{float}
\usepackage{subfigure}
\usepackage{multirow}
\usepackage{epsfig}
\usepackage{graphicx}
\usepackage{url}
\usepackage{eso-pic}
\graphicspath{{./images/}}

\wacvfinalcopy 

\def\wacvPaperID{198} 

\ifwacvfinal\fi
\begin{document}

\title{A deep learning approach to solar-irradiance forecasting in sky-videos}

\author{Talha A. Siddiqui\footnotemark[1]\\
Carnegie Mellon University\\
{\tt\small tsiddiqu@andrew.cmu.edu}
\and
Samarth Bharadwaj \\
IBM Research Labs, India\\
{\tt\small samarth.b@in.ibm.com}
\and
Shivkumar Kalyanaraman\thanks{while at IBM Research Lab, India}\\
GE Power, India \\
{\tt\small shivkumar.kalyanaraman@ge.com}
}

\maketitle
\ifwacvfinal\thispagestyle{empty}\fi
	
\begin{abstract}
Ahead-of-time forecasting of incident solar-irradiance on a panel is indicative of expected energy yield and is essential for efficient grid distribution and planning. Traditionally, these forecasts are based on meteorological physics models whose parameters are tuned by coarse-grained radiometric tiles sensed from geo-satellites. This research presents a novel application of deep neural network approach to observe and estimate short-term weather effects from videos. Specifically, we use time-lapsed videos (sky-videos) obtained from upward facing wide-lensed cameras (sky-cameras) to directly estimate and forecast solar irradiance. We introduce and present results on two large publicly available datasets obtained from weather stations in two regions of North America using relatively inexpensive optical hardware. These datasets contain over a million images that span for 1 and 12 years respectively, the largest such collection to our knowledge. Compared to satellite based approaches, the proposed deep learning approach significantly reduces the normalized mean-absolute-percentage error for both nowcasting, i.e. prediction of the solar irradiance at the instance the frame is captured, as well as forecasting, ahead-of-time irradiance prediction for a duration for upto 4 hours.
	\end{abstract}

\section{Introduction}
Long-term forecasting weather phenomenon is a challenging problem due to the vagaries of nature and extremely complex physical causation that are difficult to model accurately. However, short-term weather forecasting is a more tractable objective that can be deployed with automated correction systems (such as IoT enabled systems) to provide economic benefits. Incident solar irradiance on a surface is an important parameter that is affected by the vagaries of weather. Accurately estimating the amount of solar irradiance lends to better production estimates from solar panels. These estimates can then be used to plan for energy storage, solar automated and manual panel tracking systems \cite{meshramcost}~\cite{smartenvconf}~\cite{Singh:2017}, and panel maintenance, among other uses in alternative energy domain. In large solar farms (upwards of 1000 GW), solar energy yield forecasting is also utilized for expected yield reporting to the power grid, with monitory penalties for both under and over production. Efficient yield prediction can also improve the energy market by streamlining distribution by better matching supply with demand, drastically reducing losses and costs. 

	\begin{figure}[!t]
		\centering
		{\includegraphics[width=0.5\textwidth]{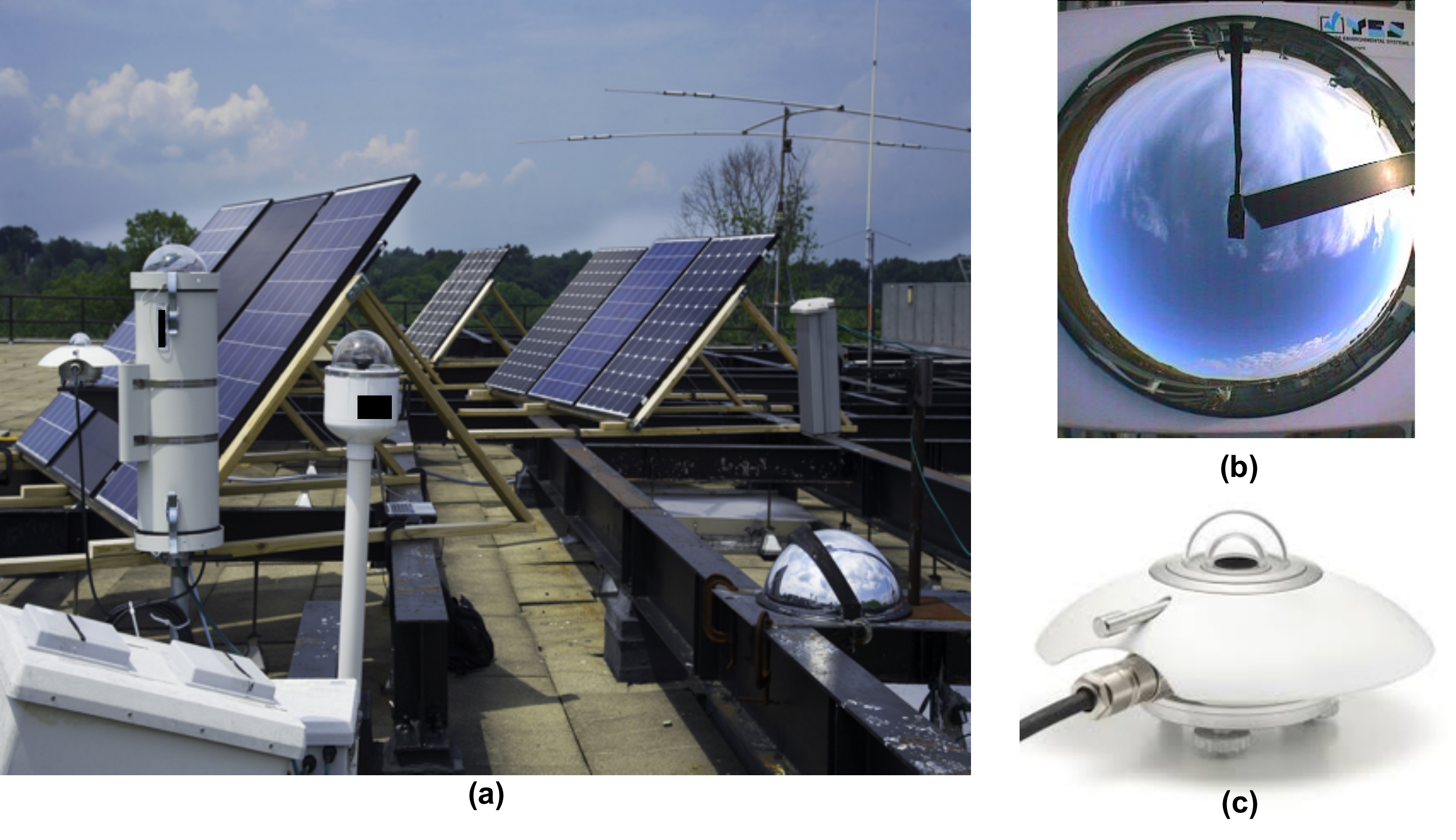}} 
		\caption{\label{fig:fig1} (a) An example of commercial sky cameras deployed in the vicinity of solar farms. (b) Sample unprocessed frame from sky-camera (TSI), (c) A thermopile pyranometer measures solar radiation flux density. While they accurately measure solar irradiance, there are no indicative measurements to forecasts. This work explores utilizing image analysis for forecasting irradiance up to 4 hours ahead-of-time.
		}
	\end{figure}
	
Solar irradiance on a surface is ideally proportional to the incident sun rays, however atmospheric elements, primarily clouds and other suspected particles can occlude, reflect, refract or diffuse sun rays in complex ways \cite{dev2017cloud}. A large cloud can be viewed as an advective fluid which has variable density and shape, that may not always exhibit laminar flow. From the camera plane, cloud motion can be approximated as a dense fluid motion along the wind direction and also simultaneously towards the camera perpendicular. Further, clouds combine and bifurcate based on turbulence and ambient weather conditions. Effectively tracking the portion of the cloud which is currently rigid using an image without explicit correspondence training samples is not well studied in literature. As described below, existing approaches rely on either colour consistency assumption or rigid flow assumption to estimate cloud behaviour. 

\begin{figure*}[!ht]
\centering
\includegraphics[width=0.99\linewidth]{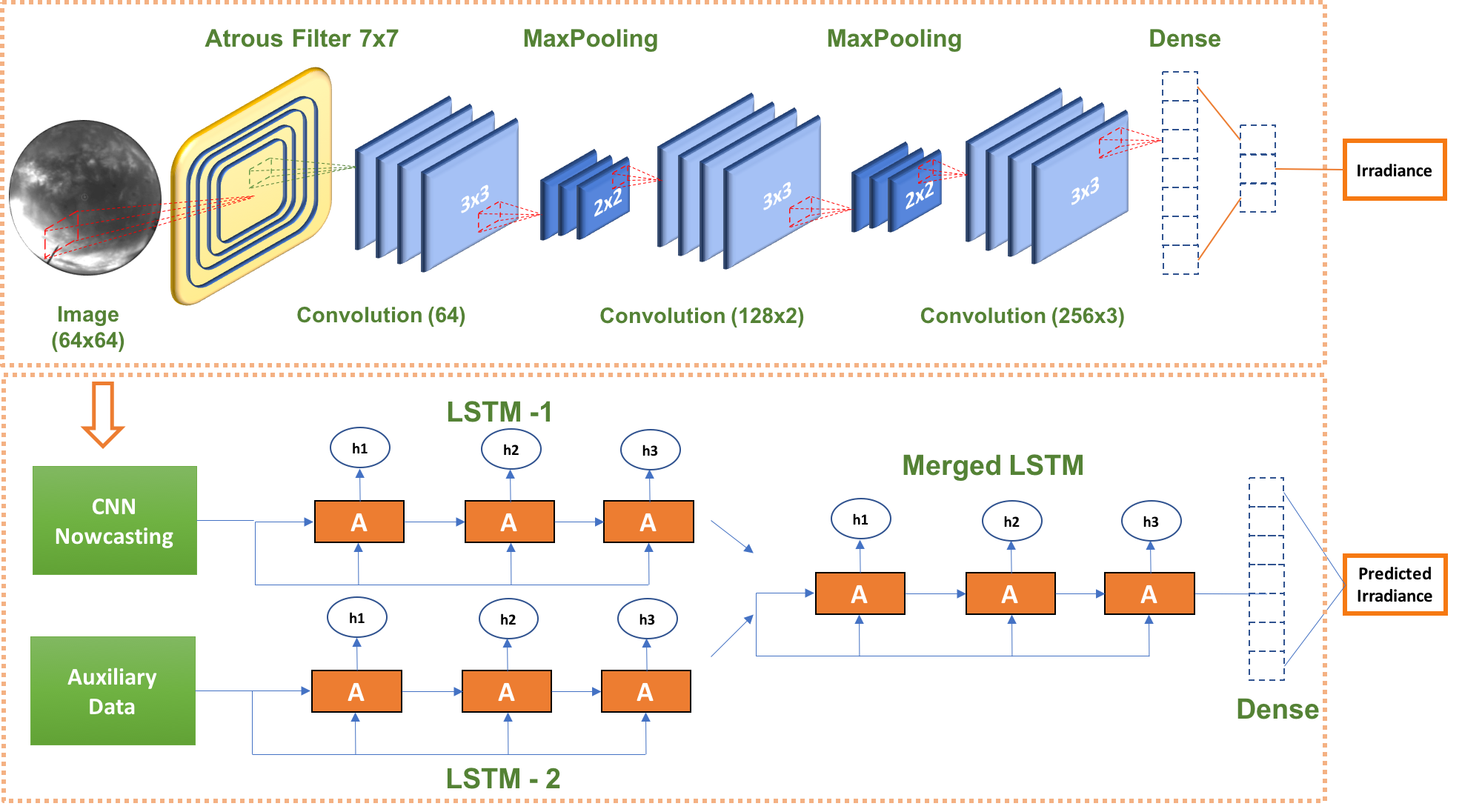}
\caption{\label{fig:prop}: Our proposed CNN+LSTM based architecture uses auxiliary weather information to guide the training process to produce short-term (between one and four hours ahead) forecasting of solar irradiance.}
\end{figure*}

\subsection{Related Work}
The earliest methods for weather prediction were geometry based models with strong assumptions on functional dependency of position, time and location~\cite{hor}, and are still used in practice. More accurate weather prediction models are based on coarse grained simulation of physical weather systems\cite{gfs}. However, such complex systems have a systematic bias to certain location, time, weather phenomenon, or unpredictable weather occurrences. In addition, the computational complexity prohibits the predictions from being of any real-time value. Further, deployment in a new location requires an elaborate re-training/fine-tuning process, traded off against error tolerance. Recently, ensemble based techniques have been introduced that combine pre-trained physics models with data driven models to fine-tune predictions \cite{lu15}. While such approaches show substantial improvement in forecast accuracy, they are limited by the availability of satellite data (typical satellite sweep ranges between 3-24hrs), need for enormous computational infrastructure, and inability to perform short-term corrections to predictions. 
	
Certain weather parameters can be predicted in short-term horizons with suitable local sensor deployment in a region. Solar-irradiance is one such measure that can be sensed with varying degrees of accuracy. Achleitner \etal ~\cite{achleitner2014sips} present an approach to aggregate several small photo-sensors for predicting irradiance. While Aryaputera \etal ~\cite{Aryaputera20151266} present a regression approach to extrapolate weather information to unknown locations. Su \etal ~\cite{su2015local} present a local feature approach to explicitly segment and track each cloud with an adaptive gaussian mixture model approach, followed by hand-crafted features for matching clouds across frames for tracking. Other pixel clustering and segmentation based approaches ~\cite{wacker2015cloud}~\cite{heinle2010automatic} explicitly measure the cloud cover from sky-images in terms of meteorological unit of \emph{okta} (the number of eighths of the sky occluded by clouds) and cloud type. Paoli \etal \cite{paoli2010forecasting} present a preliminary approach to forecasting with a shallow neural network. 

Advances in deep learning can also be leveraged for weather forecasting. Klien \etal \cite{klein2015dynamic} present a dynamic convolution approach to predict short term weather from radar imaging. Xingjian \etal \cite{Shi:2015} present a fully-connected recurrent neural network approach to short-term precipitation nowcasting. In order to forecast solar irradiance accurately, the possible occlusions from cloud cover must be tracked. Optical flow based approaches to track rigid objects have been extensively studied. Recent advances in deep learning have also been utilized for object tracking.  Weinzaepfel \etal \cite{weinzaepfel2013deepflow} presents a deep learning approach to predict correspondence images. However, these approaches require correspondence maps for training. A similar study has been done in short-term wind power forecasting. In the proposed approach by Chen \etal ~\cite{windenrgy}, Gaussian Processes applied to the outputs of a Numerical Weather Prediction model were used to perform one-day-ahead wind power forecasts. Palani \etal ~\cite{bluesky} showcase that clear sky models can be improved using a data-driven methodology and the generated model is more accurate spatio-temporally compared to the state of the art. Forecasting solar irradiance can also be utilized to re-configure solar panels accordingly. Rust \etal ~\cite{smartenvconf} demonstrate that their approach can be used to self-configure the state of smart devices in an energy-efficient manner.

\subsection{Contributions}
Solar irradiance can be measured with reasonable accuracy and high frequency using sensors such as, thermophile photo-sensors deployed locally. However, such sensors do not collect any reliable evidence of local weather phenomenon that can aid near-time forecasting. Whereas, time-lapse video (termed sky-videos) obtained from sky-camera (example deployment shown in Figure \ref{fig:fig1})  encode both incident light and atmospheric behaviour such as cloud and suspended particles. In this research, we present a two part deep neural network architecture for nowcasting and forecasting solar irradiance based on such sky-video images. A nowcast is defined as the prediction of the solar irradiance at the instance the frame is captured while a forecast is generation of ahead-of-time prediction for a duration that is updated at regular intervals. The key novelty of this work can be summarized as follows:

	\begin{itemize}
		\item
		The proposed approach uses dilating convolution filters to encode the input sky image. The resultant neural network samples the image at a higher perceptive field than a traditional neural network with fewer parameters. 
		\item
		We propose a method to utilize auxiliary data (indicative weather data such as air temperature, wind speed, relative humidity, barometric pressure) to improve the generalizability of the model on unseen data. The resultant intermediary representations provides better nowcasting results. 
		
		\item
		The approach also forms a more stable representation vector of an input image to the forecasting model. Forecasting of solar irradiance along with auxiliary data is performed with a two-tier LSTM~\cite{hochreiter1997long} architecture for up to 4 hours ahead of time.
		
		\item
		The approach is evaluated on two datasets from different locations with over a million samples that we introduce to the community. The approach out performs state-of-the-art satellite based forecasting methodology currently in use today. 
	\end{itemize}

\begin{figure*}[!ht]
\begin{center}
	\includegraphics[width=.8\textwidth]{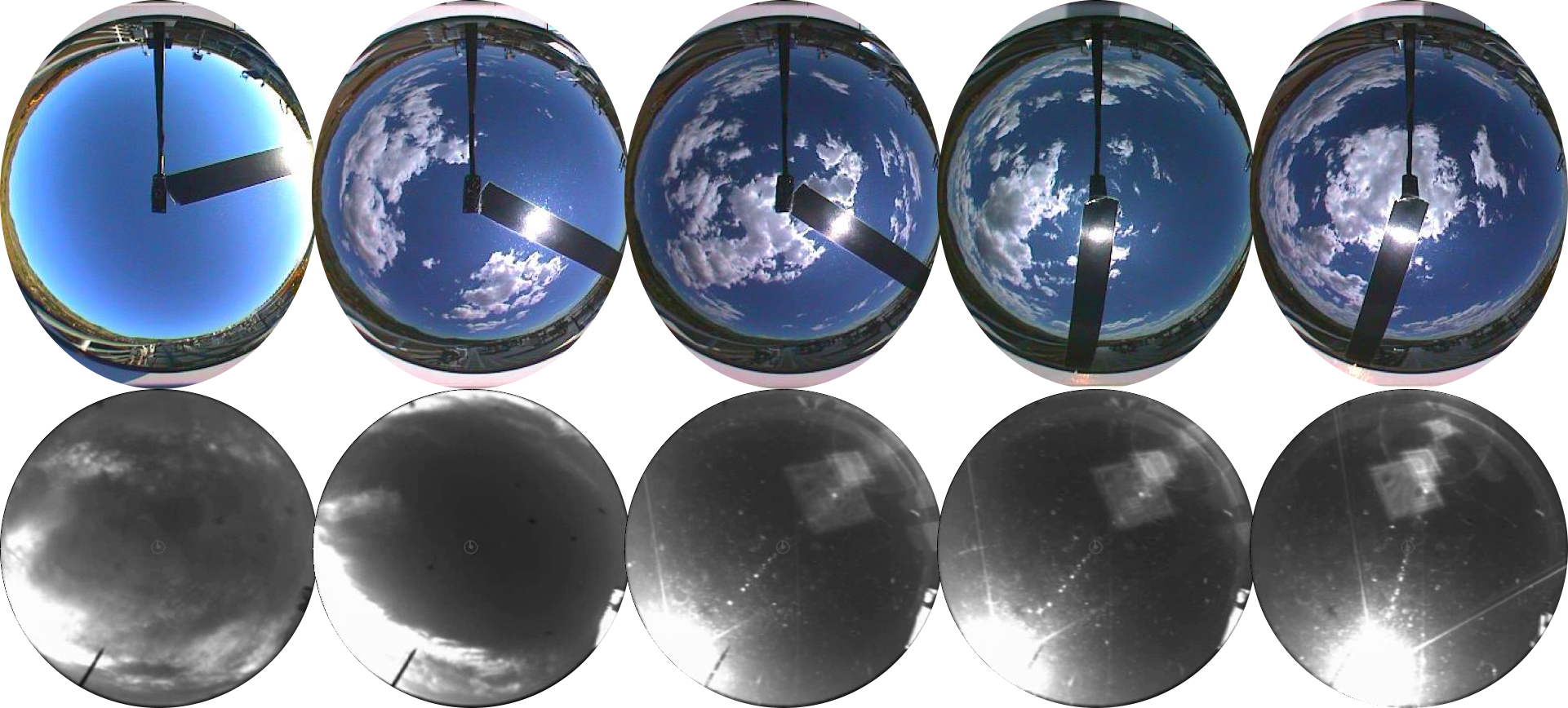}
\end{center}
\caption{Overview image of Colorado dataset (top) and Arizona dataset (bottom).}
\label{fig:overviewSRRL}
\end{figure*}

	\section{Model}
	
	In this section we describe the details and intuition of the proposed architecture for both nowcasting and forecasting prediction of solar irradiance with sky-videos, also illustrated in Figure \ref{fig:prop}. 
	
	\subsection{Architecture}
	
The proposed architecture consists of two stages: (i) A convolutional neural network stage to encode a frame from a sky-video to obtain a full-sky representation aided by auxiliary weather data, (ii) A two-tier LSTM architecture to observe historical full-sky representations and produce ahead-of-time forecasts. 
	
	\begin{itemize}
		\item
		First, a deep convolution layer is presented to encode the sky image, starting with 128 \emph{dilating} filters \cite{yu2015multi} with the initial filter size of $7\times 7$ and the dilation rate of $4\times 4$. The resultant filter has a receptive field of size $25\times 25$ which enables sampling the image with multi-scaling. 
		
		\item
		The dilation layer is followed by 64, $3\times 3$ convolution filters and reduced by a max-pooling layer of size $2\times 2$ with stride (2,2). The model then has two convolution layers of 128 filters, and a further three layers of 256 filters, all of size $3\times 3$. Each layer is followed by a $2\times 2$ max-pooling layer. The encoding features are then reduced to a 512 sized vector.
		
		\item
		The model architecture is obtained by performing ablation of layers from the original VGG$16$ \cite{vgg16} architecture. We observe that the performance improves with fewer blocks when augmented with the dilating (also known as atrous) layer. We attribute this to lower complexity of the object of interest compared to Imagenet. The model is trained with random initialization. 
		
		\item
		The auxiliary weather data, in this case a single 7 dimensional input vector, is concatenated with the 512 vector dense layer resulting in a vector of size 519. Finally, this vector is connected to a fully-connected layer along with a dropout to predict a single  unit of solar irradiance. We use the Adam \cite{kingma2014adam} model optimizer with Huber loss error (Eq. \ref{eq:huberloss}) and $L_2$-norm regularizer with suitably decaying learning rate. 
		
		\begin{equation}
            \label{eq:huberloss}
            \mathcal{H}(r, \hat{r}) := \log( m * \cosh(r - \hat{r}))
        \end{equation}
		
		\item
		In stage-2, the forecasting architecture utilizes the model from stage-1 to encode individual frames of a short \emph{look\-back} duration to obtain their corresponding full-sky representations. Next, a long short term memory recurrent neural network (LSTM) is used to learn a 128 vector representation of the historical frames. 
		
		\item
		A second LSTM simultaneously uses a 7 length weather input to produce a 4 length vector representing the corresponding auxiliary weather parameters. Both representations are then concatenated into a LSTM layer. Finally, a fully-connected layer with dropout is used for obtaining a single vector with ahead-of-time forecasting.  
		
		\item
		For the nowcasting experiment, input to the model is a single image and output is a single scalar quantity i.e. solar irradiance predicted for the given frame. For the forecasting experiment, the input is a sequence of images for a 4-hour duration and output is a vector consisting of solar irradiance for the next 4-hour frames.
	\end{itemize}
	
	\subsection{Intuition for the approach}
	
	\noindent {\bf Implicit tracking with dilating convolutions}: The problem of forecasting solar irradiance can be viewed as estimating the trajectory of clouds occluding the sky. The fluid state of a cloud with evolving shape is difficult to capture with a single convolution view. However, we assert that \emph{dilating} convolutions \cite{yu2015multi} allow us the freedom to explore wider area with similar features. Dilated convolutions are used for multi-scaling the image without increasing the number of layers in the architecture. This is pertinent for sky images with cloud movement where the receptive area needs to be rescaled to extract relevant features from an image.  Hence, a larger receptive area in the same image can be covered with lower complexity. As shown in Eq. \ref{eq:dilations},
	
	\begin{equation}
	\label{eq:dilations}
	(F*_\textit{l}k)(p)= \sum_{s+lt=p}F(s)k(t)
	\end{equation} 
	
\noindent where, *$_l$ denotes a dilated convolution between a signal $F$ and a kernel $k$. In normal convolution layers, \textit{l} is equal to 1. Therefore for a convolution filter of size $7\times 7$, a ($4,4$) dilation allows the filter to reach the receptive field of size $25\times 25$ while restricting the number of convolutions. This is a useful trade-off for sky-images as the image lacks finer or complex details such as those in typical scene images, where covering a higher receptive region of the image is more important.
	
\noindent {\bf Training with auxiliary information} sensed simultaneously with every sky-video frame using low-cost sensors helps improve model robustness. In our experiments, we use average wind speed, barometric pressure, relative humidity, and air temperature, which are known to correlate with atmospheric phenomenon that affect total incident irradiance from the sun on the panel surface. Additionally, we add Azimuth angle of the sun, derived geometrically as a function of geo-coordinate location of the camera and timestamp. Further, the estimated solar irradiance from a clearsky model is also used which measures the solar irradiance as a Cosine function of the sun's Azimuth angle $(z)$  under the assumption of cloudless skies \cite{hor}. It is given by,
	
	\begin{equation}
	\label{eq:clearsky}
	ClearSky = 1095 * \cos(z) * \exp \left ( \frac{-0.057}{\cos(z)} \right )
	\end{equation}
	\noindent 
	We hypothesize that including the auxiliary information induces the periodic nature of solar irradiance. Further, the resultant full-sky representation encodes weather and sky properties in addition to image characteristics leading to better forecast. 

	\begin{figure*}[!t] \small
		\centering
		\subfigure[\small{Tracking}]{\includegraphics[width=0.2\textwidth]{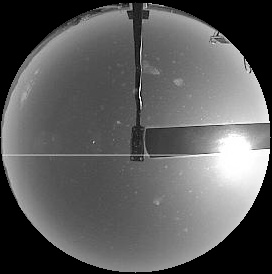}} 
		\subfigure[\small{Missed track}]{\includegraphics[width=0.2\textwidth]{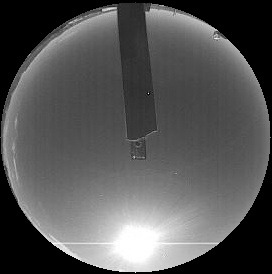}} 
		\subfigure[\small{Rain}]{\includegraphics[width=0.2\textwidth]{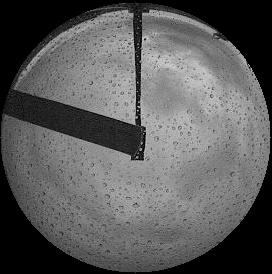}} 
		\subfigure[\small{Dew}]{\includegraphics[width=0.2\textwidth]{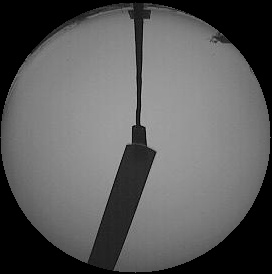}} \\
		
		\subfigure[\small{Specular reflectance}]{\includegraphics[width=0.2\textwidth]{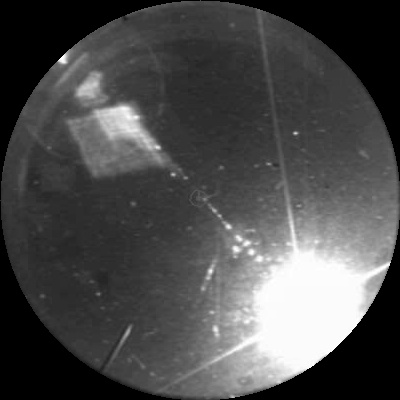}} 
		\subfigure[\small{ Rain drop or dew}]{\includegraphics[width=0.2\textwidth]{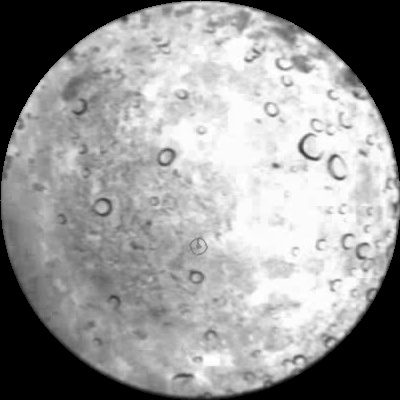}} 
		\subfigure[\small{Specular reflectance}]{\includegraphics[width=0.2\textwidth]{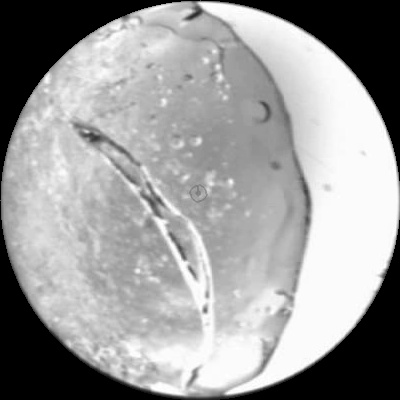}} 
		\subfigure[\small{ Rain drop or dew}]{\includegraphics[width=0.2\textwidth]{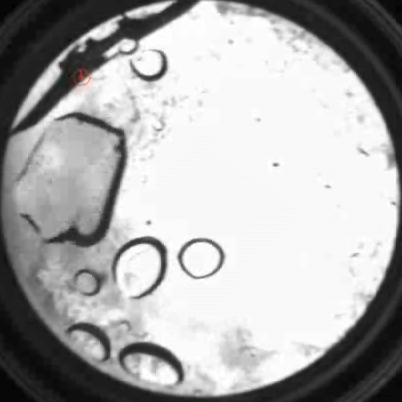}} 
		\caption{\label{fig:chal} Challenging skycam frames from both datasets (top: Colorado, bottom: Arizona) after pre-processing to remove bezel. We do not redact any image from the dataset to emulate real-world conditions.}
	\end{figure*}	
\section{Evaluation}
	
A sky-video is obtained from an upward facing wide-angle lensed video camera such as the one shown in Figure \ref{fig:fig1}. The obtained images are circular and sample the full-sky region (illustrated in Figure \ref{fig:overviewSRRL}). Based on the configuration of the camera, the sun region may be occluded to prevent image saturation or cameras with anti-blooming filters may be used.  The input to the architecture is a normalized image of dimensions $(64, 64, 3)$.  
	
\subsection{Sky-video datasets}
We showcase the performance of our approach on two publicly available datasets of sky-videos obtained from two different locations in the United States\footnote{the datasets are available to download freely on the corresponding websites.}. Each of the two datasets use different cameras to record the videos. The description for each of the datasets is as follows:
	
	\noindent {\bf Golden, Colorado Dataset}: is recorded at Solar Radiation Research Laboratory (SRRL) \cite{srrl}. Colorado is situated in North America, surrounded with Rocky Mountains and receives high rains during July and August. The dataset has been recorded using a commercial camera (TSI) \cite{morris2005total} that provides a wide angle view of the sky and records frames at every 10 minutes interval. A mechanical sun tracker is used to block the sun to prevent saturation in the image. The dataset is available for the last 12 years from 2004-2016 and the total images captured are 304,309.  
	
	\noindent  {\bf Tuscan, Arizona Dataset}: has been recorded at the Multiple Mirror Telescope Observatory (MMTO) \cite{mmto}. Arizona is located in south-west region of North America and observes majorly two seasons - Summer and Winter. Arizona dataset was created by an in-house camera developed at MMTO \cite{mmto} with custom hardware specifications that captures both RGB and near infrared light. It is a low cost sky camera with a wide angle view of about 150 degrees  of the sky and blooming filter to prevent over-saturation. The camera records approximately 10 frames per minute giving us a finer representation of the changes in the sky. The dataset is available from the months November, 2015 to May, 2016. For our experimental purpose we are using images only from sunrise to sunset and approximately one million images (993,101) are recorded during this period.

\noindent The wide-angle images from the fish-eye lens are used with suitable padding without any rectification due to the unavailability of camera calibration matrices. Figure \ref{fig:chal} shows some challenging examples images. Apart from the frames, we also use auxiliary weather data obtained from nearby deployed sensors. 
		
\subsection{Protocol}
	
Two types of predictions are performed on the datasets, namely Nowcasting and Forecasting and the below section describes the experimental protocol for both.
	
	\noindent {\bf Nowcasting}: The Arizona dataset contains images from November, 2015 to May, 2016. We have split our dataset such that images from November to February are used in training and from March to May are used in testing. Total images in training and testing sets are 524,272 and 468,829 respectively. This experimental protocol mimics a practical deployment scenario where historical data is used to tune the system for future predictions. 
	
	The Colorado dataset has images captured over 12 years from 2004-2016. Since it has a wide range of data, we have trained a model on the first 10 years i.e. from 2004-2014 and tested the model on 2015-2016 data. Total images in training and testing are 251,600 and 52,709 respectively. Experiments with and without the aid auxiliary data are performed.
	
	\noindent {\bf Forecasting}:  For the Arizona dataset, we are using 7 months of data to train and test our model. The split between train and test data is same as in the nowcasting experiment. However since the dataset is huge, we are using every 4th frame for the experiment. We are taking four hour previous images and weather data to look back and predict the next 4 hours solar irradiance. We are taking 2 images in a minute, creating a single training sample consisting of 480 images for four hour look-back. Each new training sample is created in every half an hour i.e. every half hour, the new data is pushed in the training sample and the latter half hour data is pushed out. There are a total of 2554 training sample generated from the train data. 
	
For Golden, Colorado dataset, the train and test split is also same as the nowcasting experiment. We have used a look-back of 6 hours to predict the solar irradiance generated in the next 4 hours. Since the total number of images captured in one hour are 6 (1 in every 10 minutes), total images in one training sample are 36 along with their corresponding weather data. Each new training sample is created at an interval of one hour i.e. after every one hour, the latter one hour data is pushed out and the latest one hour data is pushed in to create a new training sample. Since the Colorado data is recorded for a longer duration, total training samples generated are 31,005. Predictions are given for the next 4 hours in an interval of 10 minutes (same as the frequency of the dataset).
	
	\begin{table}[!ht] 
	\footnotesize
		\begin{center}
			\caption{\label{tab:now} \small nMAP Error for nowcasting using different techniques on multiple databases. The performance of first three blocks of VGG16 is better than the full VGG16 model when using dilation filters. (ww- without weather)}
			\begin{tabular}{c|c c| c c c }
				\multirow{2}{*}{Experiment} & \multicolumn{2}{c|}{Colorado} & \multicolumn{3}{c}{Arizona} \\\cline{2-6}
				& \multicolumn{1}{c}{2015} & \multicolumn{1}{c|}{2016} & \multicolumn{1}{c}{March} & \multicolumn{1}{c}{April} & \multicolumn{1}{c}{May} \\\hline
				{\small VGG16 (rand init)} & 21.0 & 21.9 &14.3 & {22.8} & 24.9    \\
				Ours (ww) & 15.9 & 17.3 & 24.4 & 58.5 & 77.4   \\
				Ours & {\bf 14.6} & {\bf 15.7} & {\bf 11.4} & {\bf 20.7} & {\bf21.4}   \\
				Auxiliary only & 31.9 & 35.3 & 26.5 & 29.9 & 23.5 \\
			\end{tabular}
		\end{center}
	\end{table}
	
	\begin{table*}[!ht] \footnotesize
		\begin{center}
			\caption{\label{tab:foreSRRL} \small nMAP Error for forecasting on Colorado database.}
			\begin{tabular}{c|c c c c|c c c c}
				\multirow{3}{*}{Experiment} & \multicolumn{8}{c}{Colorado} \\\cline{2-9}
				& \multicolumn{4}{c|}{2015} & \multicolumn{4}{c}{2016}\\\cline{2-9}
				& \multicolumn{1}{c}{+1hr} & \multicolumn{1}{c}{+2hr} & \multicolumn{1}{c}{+3hr} & \multicolumn{1}{c|}{+4hr} & \multicolumn{1}{c}{+1hr} & \multicolumn{1}{c}{+2hr} & \multicolumn{1}{c}{+3hr} & \multicolumn{1}{c}{+4hr}\\\hline
				Ours &17.9  &25.2  &{\bf 31.6}  &39.1 &16.9  &25.0  &{\bf 31.9}  &39.5 \\
				ECMWF & - & - & - & - & - & -  & 77.6 & - \\
				GFS & - & - & 110.5 & - & - & -  & 115.8 & - \\
				Auxiliary only & 41.7 & 44.3 & 45.2 & 46.3 & 45.5 & 47.3  & 48.5 & 50.2 \\
			\end{tabular}        
		\end{center}
	\end{table*}
	
	\begin{table*} \footnotesize
		\begin{center}
			\caption{\label{tab:foreMMTO} \small nMAP Error for forecasting for Arizona database.}
			\begin{tabular}{c|c c c c|c c c c|c c c c}
				\multirow{3}{*}{Experiment} & \multicolumn{12}{c}{Arizona} \\\cline{2-13}
				& \multicolumn{4}{c|}{March} & \multicolumn{4}{c|}{April}& \multicolumn{4}{c}{May}\\\cline{2-13}
				& \multicolumn{1}{c}{+1hr} & \multicolumn{1}{c}{+2hr} & \multicolumn{1}{c}{+3hr} & \multicolumn{1}{c|}{+4hr} & \multicolumn{1}{c}{+1hr} & \multicolumn{1}{c}{+2hr} & \multicolumn{1}{c}{+3hr} & \multicolumn{1}{c|}{+4hr} & \multicolumn{1}{c}{+1hr} & \multicolumn{1}{c}{+2hr} & \multicolumn{1}{c}{+3hr} & \multicolumn{1}{c}{+4hr}\\\hline
				Ours  & 29.7 &  33.1 & {\bf 34.5}& 28.8 & 49.7 & 52.8 & {\bf 52.3} & 40.6 & 56.1 & 55.6 & {\bf 53.7} & 47.0\\                    
				GFS & - & - & 96.0 & - & - & - & 107.8 & - & - & - & 111.9& -\\
				Auxiliary only & 53.5 & 62.8 & 44.3 & 53.6 & 65.3 & 71.8 & 56.9 & 63.7 & 58.0 & 66.4 & 54.3 & 57.9\\
				
			\end{tabular}
		\end{center}
	\end{table*}

	
\subsection{Analysis}
	
The results are reported as normalized mean absolute percentage (nMAP) error of predictions, given by Equation \ref{eq:nmape}. Nowcasting and forecasting errors are summarized in Table \ref{tab:now} and Table \ref{tab:foreSRRL} \& \ref{tab:foreMMTO} respectively.

\begin{equation}
\label{eq:nmape}
nMAP = \frac{1}{n}\sum_{i=1}^{n} \frac{ |r_i - \hat{r_i}|}{\frac{1}{n}\sum_{i=1}^{n}(r_i)} \times 100
\end{equation} 
	
\noindent where $r_i$ and $\hat{r_i}$ are ground truth and predicted irradiance respectively. We demonstrate the performance of our proposed approach along with traditional VGG$16$ deep learning framework \cite{vgg16} on both the datasets with random weight initialization. 
	
\begin{figure}
\begin{center}
	\subfigure[\small{Colorado: Forecasting}]  { \includegraphics[width=0.45\textwidth]{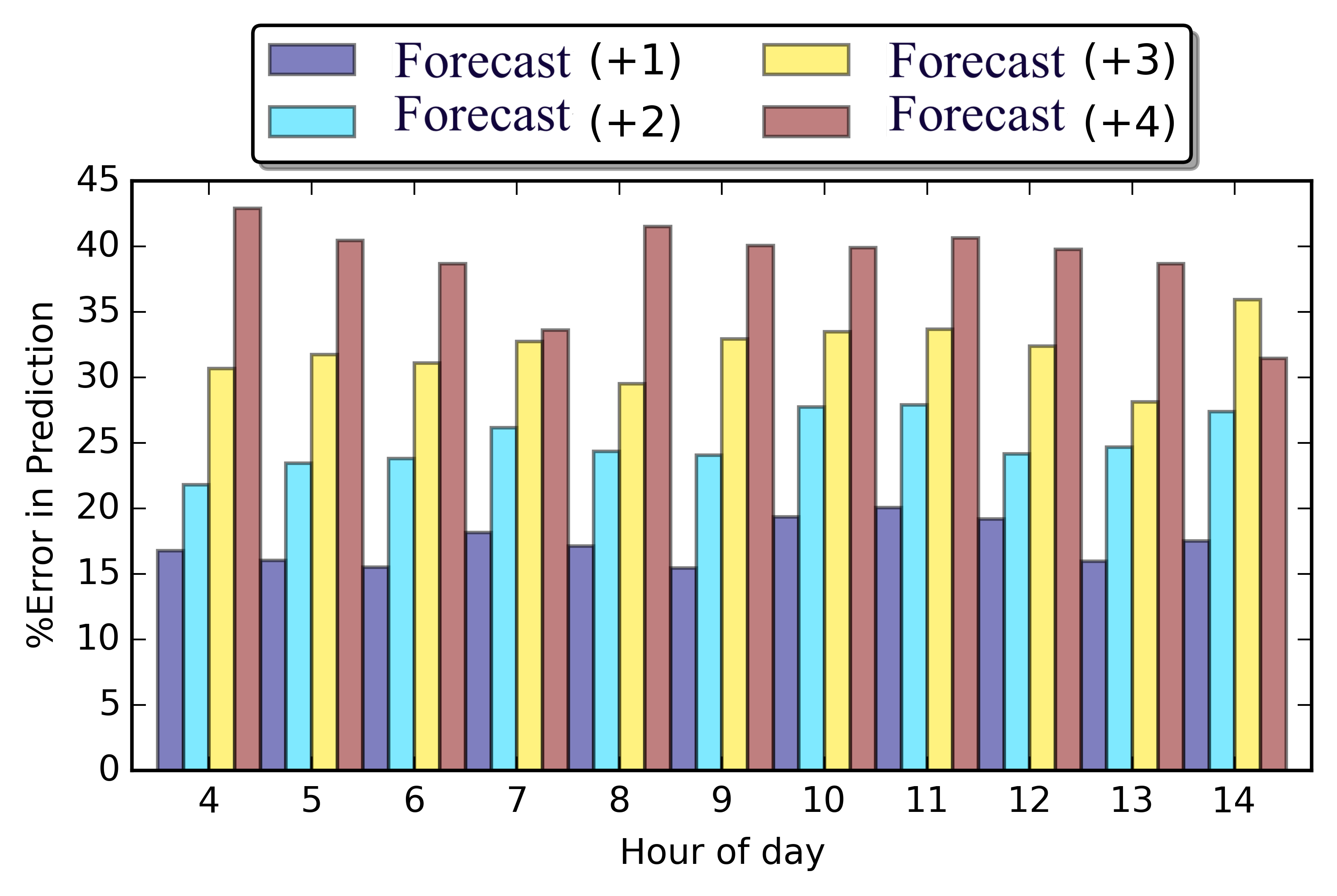}} 
	\subfigure[\small{Arizona: Forecasting}]{\includegraphics[width=0.45\textwidth]{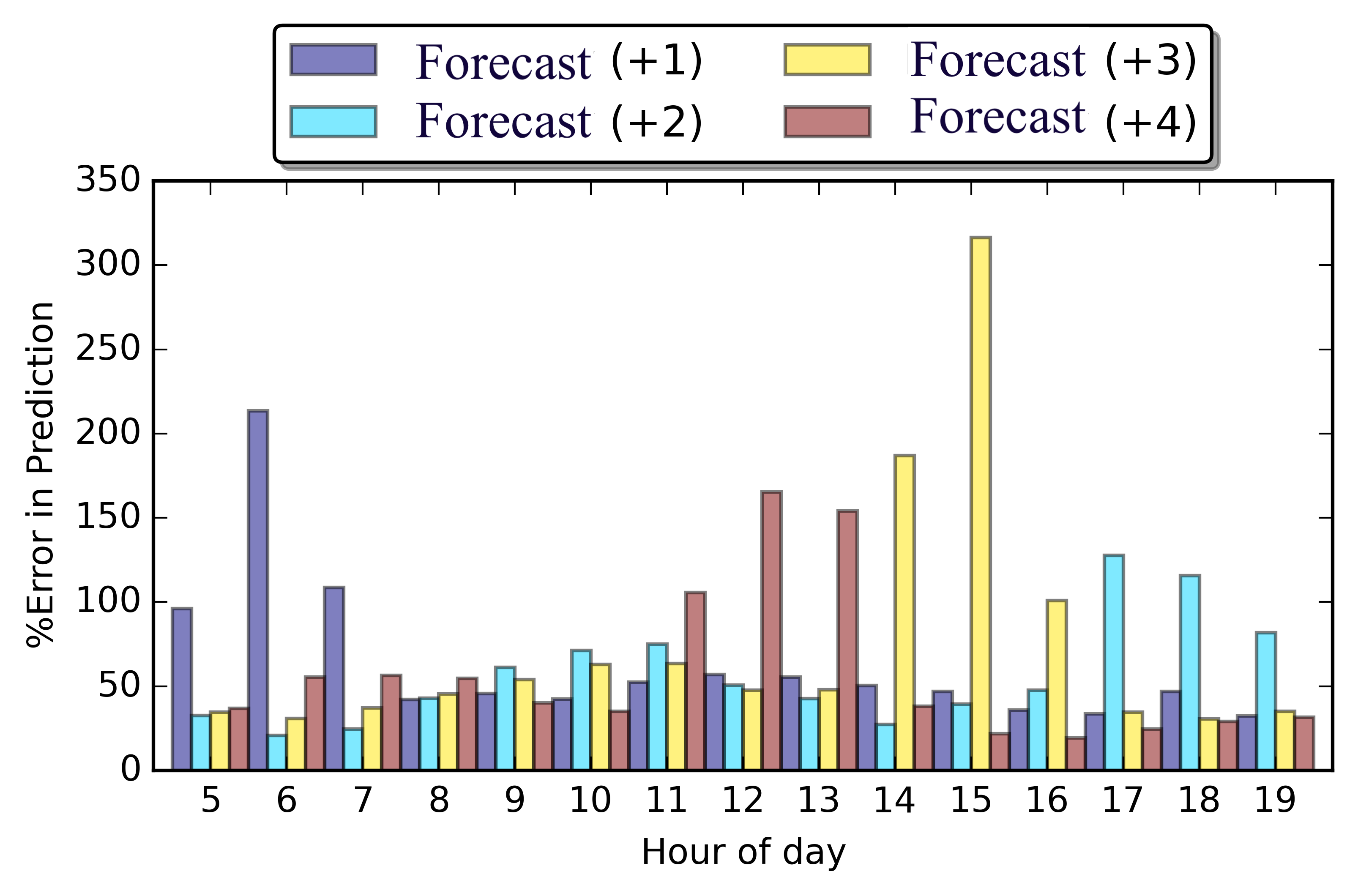}}
\end{center}
\caption{Hourly nMAP in ahead-of-time forecast of +1, +2, +3, and +4 hours. (Best viewed in colour)}
\label{fig:forecast}
\end{figure}

\begin{figure*}
\begin{center}
	\includegraphics[width=0.90\linewidth]{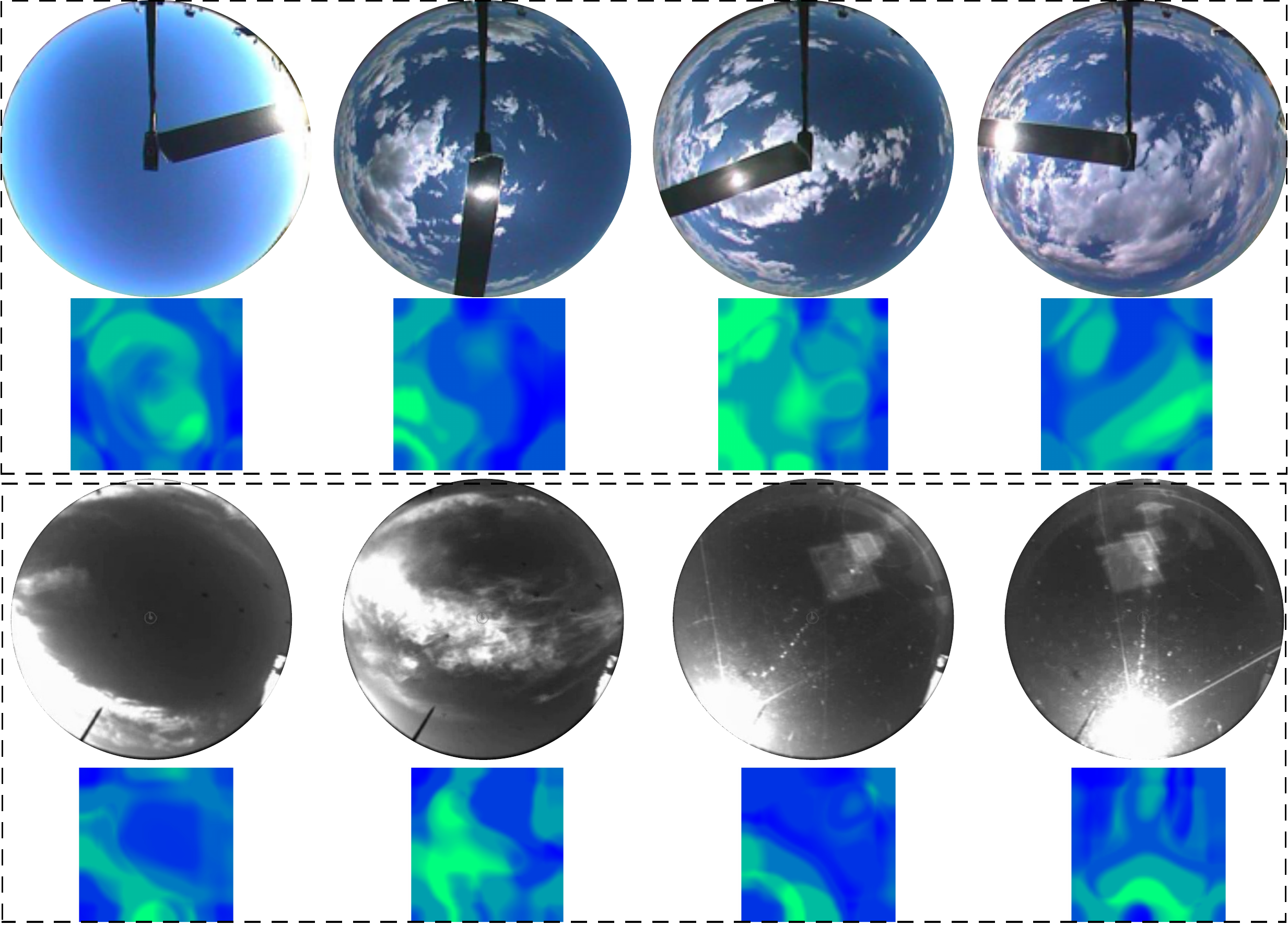}
\end{center}
\caption{Consecutive frames from the Colorado dataset (top) and Arizona dataset (bottom) plotted with interpolated mean of the \emph{hypercolumns} indicative of focus of CNNs. The heat maps indicate our model's ability to capture flow in the video (better visualized in the supplementary material in video format.).}
\label{fig:timelag}
\end{figure*}

\noindent Our experiments on two publicly available datasets with very different camera characteristics and deployment locations prove the generalizability of our  approach. We avoid data augmentation by orientation or zooming, as they could change the position of sun with respect to the camera.

Consistent protocol is also used for comparison with publicly available Global Forecast System (GFS) \cite{gfs} deployed by the National Weather Services, USA. GFS is a state-of-the-art numerical weather prediction (Flow-following, finite-volume icosahedral) models that produces 4 forecasts a day. These models are computational intensive and provide sparse daily updates due to latency in availability of satellite image data. Similarly, we also compare with forecasts obtained from European Centre for Medium-Range Weather Forecasts (ECMWF), a proprietary service available to us for Colorado region in 2016.

\begin{itemize}
	\item
	{\bf Auxiliary parameters aid learning}: We augment the training of the CNNs with auxiliary weather parameters, namely, average windspeed, relative humidity, barometric pressure, air temperature, sun position ($z$), and clear sky prediction (from Eq. \ref{eq:clearsky}), and observe that the neural network architectures converge faster and achieve lower errors. As shown in Table \ref{tab:now}, nowcasting nMAP trained with weather parameters, as opposed to just sky images,  has an overall 1.46 decrease in error in Colorado, and an overall of 36.02 decrease in Arizona dataset. The large improvement in Arizona dataset can be attributed to the lower image quality (see Figure \ref{fig:overviewSRRL}) and the large temporal distance between training and testing data.
	

	\item We observe that all models struggle to predict irradiance in early morning and late evening, both for nowcasting and forecasting. Furthermore, the ground truth measurement of solar irradiance in these hours is also unreliable (affected by shadows or diffusion) and not useful for energy production (low absolute value). 
			
	\item
{\bf Dilating convolutions}: As mentioned previously, the first set of convolution filters applied to the image are dilated in size. We apply a 7$\times$7 filter with a dilation rate of  4$\times$4, thereby making the receptive field of size 25$\times$25. The size and dilation constant were selected based on empirical observation on a series of experiments performed with filter size varying between 3$\times$3 to 7$\times$7. It was observed that larger filter sizes resulted in better convergence over the training set and lower nMAP error on the test set.
	
	\item
	Figure \ref{fig:forecast} shows the mean ahead-of-time forecasting errors in hourly fashion. As expected, the forecasting error increases for larger time lapse. Note that, to obtain irradiance forecast for the beginning hours of a given day, the sky images and auxiliary data from the evening of the previous day are used, in a cyclic fashion. 
	
	\item
	Compared to existing approaches of solar irradiance prediction, the proposed approach provides comparable results and the additional flexibility of frequent forecasts. All presented forecasting experiments are conducted with prediction update at regular intervals (30 minutes for Arizona, one hour for Colorado). Although the model has smaller number of model parameters as compared to full VGG16 architecture, the model provides superior nowcasting and forecasting error rates. 
	
	\item 
	The ahead-of-time forecasting errors for irradiance, shown in Table \ref{tab:foreSRRL}, \ref{tab:foreMMTO} are obviously higher than the corresponding nowcasting errors (in Table \ref{tab:now}. However, they are also influenced by the time of day variability in weather. From an application point of view, the performance of the model between 9:00am to 3:00pm (peak average solar irradiance) is satisfactory. Further, forecasting error four hours ahead is lower due to small number of valid samples within the daylight window.  
	
	\item
	{\bf Focus of convolution filters}: We plot the interpolated mean of the \emph{hypercolumns} \cite{hariharan2015hypercolumns} obtained from all the convolution filters on each frame to aid visualization. Figure \ref{fig:timelag} shows weather phenomenon from consecutive frames of both the datasets being tracked by the convolution filters.
	
	\item
	{\bf Auxiliary only experiments}: We also perform nowcasting and forecasting prediction with a linear regression model. Specifically, the aforementioned  auxiliary parameters are used to regress to the instantaneous irradiance for nowcasting. Similarly, a regression extrapolation is predicted with  auxiliary parameters from previous hour. The regression models outperform GFS and ECMWF as it is fine-tuned to the specific location and local weather. Further, regression is less effected by monthly drift as it is only fitted over data from the previous window.   
	
	\item
    The Colorado dataset benefits from multiple years of data to capture seasonality. However, the performance on the Arizona dataset, though comparable with physics based numerical models, is achieved despite low diversity in training samples (only winter seasonal months).  It is observed that the performance degrades month-on-month as the testing data distribution drifts away from the training set, corresponding to seasonal change from winter to summer. 
\end{itemize}

	\section{Conclusion and Future Work}
	Ahead-of-time prediction of irradiance that influence production, yield, and efficiency of solar farms is critical for risk assessment and grid planning. Many national power grid agencies have began enforcing slab penalties for incorrect daily power generation commitments. This research presents the largest such study to process sequences of video frames obtained from full-sky imaging and forecast solar irradiance 1-4 hours ahead of time, using deep neural networks. In two separate locations we show that the proposed deep learning approach out-performs other solutions for nowcasting and forecasting predictions of surface irradiance at a fraction of the infrastructure cost. 
	We are currently deploying an open web based solution to aggregate video-feed from several spatially distributed sky-cameras from multiple partners to be used as a crowd-sourced prediction platform. Data and scripts will be open-sourced for easy reproducibility (\url{https://bit.ly/2Bw7HGP}).

	\clearpage
	{ \small
	\bibliographystyle{ieee}
	\bibliography{skycam}

\begin{thebibliography}{10}\itemsep=-1pt

\bibitem{achleitner2014sips}
S.~Achleitner, A.~Kamthe, T.~Liu, and A.~E. Cerpa.
\newblock Sips: Solar irradiance prediction system.
\newblock In {\em Proceedings of the 13th international symposium on
  Information processing in sensor networks}, pages 225--236. IEEE Press, 2014.

\bibitem{Aryaputera20151266}
A.~W. Aryaputera, D.~Yang, L.~Zhao, and W.~M. Walsh.
\newblock Very short-term irradiance forecasting at unobserved locations using
  spatio-temporal kriging.
\newblock {\em Solar Energy}, 122:1266 -- 1278, 2015.

\bibitem{windenrgy}
N.~Chen, Z.~Qian, I.~Nabney, and X.~Meng.
\newblock Short-term wind power forecasting using gaussian processes.
\newblock In {\em International Joint Conference on Artificial Intelligence},
  2013.

\bibitem{dev2017cloud}
S.~Dev, S.~Manandhar, F.~Yuan, Y.~H. Lee, and S.~Winkler.
\newblock Cloud radiative effect study using sky camera.
\newblock In {\em USNC-URSI Radio Science Meeting (Joint with AP-S Symposium),
  2017}, pages 65--66. IEEE, 2017.

\bibitem{hariharan2015hypercolumns}
B.~Hariharan, P.~Arbel{\'a}ez, R.~Girshick, and J.~Malik.
\newblock Hypercolumns for object segmentation and fine-grained localization.
\newblock In {\em Proceedings of the IEEE Conference on Computer Vision and
  Pattern Recognition}, pages 447--456, 2015.

\bibitem{heinle2010automatic}
A.~Heinle, A.~Macke, and A.~Srivastav.
\newblock Automatic cloud classification of whole sky images.
\newblock {\em Atmospheric Measurement Techniques}, 3(3):557, 2010.

\bibitem{hochreiter1997long}
S.~Hochreiter and J.~Schmidhuber.
\newblock Long short-term memory.
\newblock {\em Neural computation}, 9(8):1735--1780, 1997.

\bibitem{kingma2014adam}
D.~Kingma and J.~Ba.
\newblock Adam: A method for stochastic optimization.
\newblock {\em arXiv preprint arXiv:1412.6980}, 2014.

\bibitem{klein2015dynamic}
B.~Klein, L.~Wolf, and Y.~Afek.
\newblock A dynamic convolutional layer for short range weather prediction.
\newblock In {\em Proceedings of the IEEE Conference on Computer Vision and
  Pattern Recognition}, pages 4840--4848, 2015.

\bibitem{lu15}
S.~Lu, Y.~Hwang, I.~Khabibrakhmanov, F.~J. Marianno, X.~Shao, J.~Zhang, B.-M.
  Hodge, and H.~F. Hamann.
\newblock Machine learning based multi-physical-model blending for enhancing
  renewable energy forecast-improvement via situation dependent error
  correction.
\newblock In {\em Control Conference (ECC), 2015 European}, pages 283--290.
  IEEE, 2015.

\bibitem{meshramcost}
S.~Meshram, S.~Valvi, and N.~Raykar.
\newblock A cost-effective microcontroller based sensor for dual axis solar
  tracking.
\newblock {\em International Conference on Renewable Energies and Power
  Quality}, 2014.

\bibitem{morris2005total}
V.~Morris.
\newblock Total sky imager (tsi) handbook.
\newblock {\em Handbook}, 2005.

\bibitem{gfs}
G.~NOAA.
\newblock Global forecast system.
\newblock {\em National Centers for Environmental Prediction
  (\url{www.ncdc.noaa.gov})}, 2019.

\bibitem{srrl}
NREL Solar Radiation Research Laboratory (SRRL).
\newblock {\em Baseline Measurement System (BMS)
  (\url{https://midcdmz.nrel.gov/srrl_bms/})}, 1981.

\bibitem{bluesky}
K.~Palani, R.~Kota, A.~Azad, and V.~Arya.
\newblock Blue skies: A methodology for data-driven clear sky modelling.
\newblock In {\em International Joint Conference on Artificial Intelligence},
  2017.

\bibitem{paoli2010forecasting}
C.~Paoli, C.~Voyant, M.~Muselli, and M.-L. Nivet.
\newblock Forecasting of preprocessed daily solar radiation time series using
  neural networks.
\newblock {\em Solar Energy}, 84(12):2146--2160, 2010.

\bibitem{mmto}
T.~Pickering.
\newblock The mmt all-sky camera.
\newblock In {\em SPIE Astronomical Telescopes+ Instrumentation}, pages
  62671A--62671A. International Society for Optics and Photonics, 2006.

\bibitem{hor}
M.~J. Reno, C.~W. Hansen, and J.~S. Stein.
\newblock Global horizontal irradiance clear sky models: Implementation and
  analysis.
\newblock {\em Tech. Report}, 2012.

\bibitem{smartenvconf}
P.~Rust, G.~Picard, and F.~Ramparany.
\newblock Using message-passing dcop algorithms to solve energy-efficient smart
  environment configuration problems.
\newblock In {\em International Joint Conference on Artificial Intelligence},
  2016.

\bibitem{Shi:2015}
X.~Shi, Z.~Chen, H.~Wang, D.-Y. Yeung, W.-k. Wong, and W.-c. Woo.
\newblock Convolutional lstm network: A machine learning approach for
  precipitation nowcasting.
\newblock In {\em Proceedings of the 28th International Conference on Neural
  Information Processing Systems}, NIPS'15, pages 802--810, Cambridge, MA, USA,
  2015. MIT Press.

\bibitem{vgg16}
K.~Simonyan and A.~Zisserman.
\newblock Very deep convolutional networks for large-scale image recognition.
\newblock {\em arXiv preprint arXiv:1409.1556}, 2014.

\bibitem{Singh:2017}
N.~Singh, P.~Dayama, S.~Randhawa, K.~Dasgupta, M.~Padmanaban, S.~Kalyanaraman,
  and J.~Hazra.
\newblock Photonic energy harvesting: Boosting energy yield of commodity solar
  photovoltaic systems via software defined iot controls.
\newblock In {\em Proceedings of the Eighth International Conference on Future
  Energy Systems}, e-Energy, 2017.

\bibitem{su2015local}
F.~Su, W.~Jiang, J.~Zhang, H.~Wang, and M.~Zhang.
\newblock A local features-based approach to all-sky image prediction.
\newblock {\em IBM Journal of Research and Development}, 59(2/3):6--1, 2015.

\bibitem{wacker2015cloud}
S.~Wacker, J.~Gr{\"o}bner, C.~Zysset, L.~Diener, P.~Tzoumanikas,
  A.~Kazantzidis, L.~Vuilleumier, R.~St{\"o}ckli, S.~Nyeki, and N.~K{\"a}mpfer.
\newblock Cloud observations in switzerland using hemispherical sky cameras.
\newblock {\em Journal of Geophysical Research: Atmospheres}, 120(2):695--707,
  2015.

\bibitem{weinzaepfel2013deepflow}
P.~Weinzaepfel, J.~Revaud, Z.~Harchaoui, and C.~Schmid.
\newblock Deepflow: Large displacement optical flow with deep matching.
\newblock In {\em Proceedings of the IEEE International Conference on Computer
  Vision}, pages 1385--1392, 2013.

\bibitem{yu2015multi}
F.~Yu and V.~Koltun.
\newblock Multi-scale context aggregation by dilated convolutions.
\newblock {\em arXiv preprint arXiv:1511.07122}, 2015.

\end{thebibliography}
	}
	
\end{document}